# Analysis of classifier training on synthetic data for cross-domain datasets

Andoni Cortés, Clemente Rodríguez, Gorka Vélez, Javier Barandiarán, and Marcos Nieto

*Abstract*—A major challenges of deep learning (DL) is the necessity to collect huge amounts of training data. Often, the lack of a sufficiently large dataset discourages the use of DL in certain applications. Typically, acquiring the required amounts of data costs considerable time, material and effort. To mitigate this problem, the use of synthetic images combined with real data is a popular approach, widely adopted in the scientific community to effectively train various detectors. In this study, we examined the potential of synthetic data-based training in the field of intelligent transportation systems. Our focus is on camera-based traffic sign recognition applications for advanced driver assistance systems and autonomous driving. The proposed augmentation pipeline of synthetic datasets includes novel augmentation processes such as structured shadows and gaussian specular highlights. A well-known DL model was trained with different datasets to compare the performance of synthetic and real image-based trained models. Additionally, a new, detailed method to objectively compare these models is proposed. Synthetic images are generated using a semi-supervised errors-guide method which is also described. Our experiments showed that a synthetic image-based approach outperforms in most cases real image-based training when applied to cross-domain test datasets (+10% precision for GTSRB dataset) and consequently, the generalization of the model is improved decreasing the cost of acquiring images.

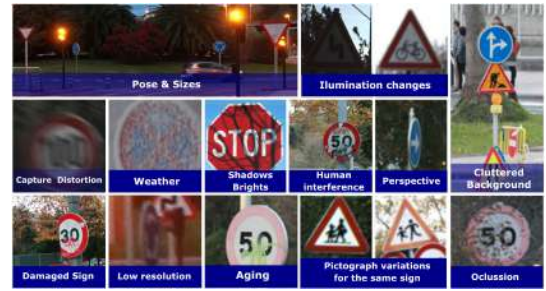

Fig. 1. Traffic signs of different appearances. This variety is a major challenge in classification problems.

## I. INTRODUCTION

TRANSPORTATION systems around the world are becoming technologically more complex every day. The number of vehicles in the transportation system is also increasing and transportation channels are getting congested. Thus, ensuring safe driving has become a major global challenge. Advanced driver assistant systems (ADAS) are a major step toward achieving safe driving because they (i) support the driver by providing assistance in fundamental skills and (ii) serve as end-to-end systems that support self-driving vehicles. These trends are reflected in the automobile market as the automotive industry is regularly introducing ADAS in new vehicles to minimize accidents and thus increase safety. ADAS include blind spot detection (BSD), lane departure warning (LDW), collision avoidance systems (CAS), traffic sign recognition (TSR), driver drowsiness detectors (DDD), etc. For example, TSR systems analyze images of a forward looking camera to detect and recognize traffic signs. This can be used either to inform or warn the driver about relevant current traffic rules or to feed the decision system of an autonomous driving system.

The emergence and evolution of deep learning (DL) have delivered remarkable breakthroughs in object classification and detection systems, reaching levels of human reliability. However, DL techniques are heavily influenced by the quality, diversity and volume of training datasets. In this safety-related context, model accuracy is required to overcome the inherent challenges of computer vision (Fig. 1).

The contributions of this paper are discussed thus. First, we present an in-depth analysis of the usage of synthetic data alone in the learning process of a simple convolutional neural network (CNN) classification model, applied to TSR. Second, we propose a new method to objectively compare classifiers when different heterogeneous datasets are available, that is, when datasets have different cardinality and traffic sign types.

Third, a proposal of facing for classification model training is presented, following the guidelines obtained from the empirical analysis.

This remaining part of this paper is structured as follows. Section 2 is a general overview of related works, describing the methodologies used and some significant results. In Section 3, the complete process of synthetic dataset generation is described. Section 4 focuses on a detailed explanation of the analysis, deepening into the methods used for the comparison between models. Finally, Section 5 presents the analysis results, discussion and conclusions.

## II. RELATED WORK

**Advanced driving asistance systems** (ADAS) are core systems that will lead the conduction towards autonomous vehicles scenarios in the near future. There is extensive literature on these systems, reviews [1]–[3]. In this paper, we focus on TSR systems which are a major element required

A. Cortés, G. Vélez, J. Barandiarán, and M. Nieto are with the Deparment of Intelligent Transportation Systems and Engineering Deparment, Vicomtech, Paseo Mikeletegi 57, San Sebastian, 20009,Spain, e-mail: {acortes, gvelez, mnieto, jbarandiaran}@vicomtech.org

C. Rodríguez is with the Universidad del País Vasco (UPV), Dpt. of Computer Architecture, San Sebastian (Spain)

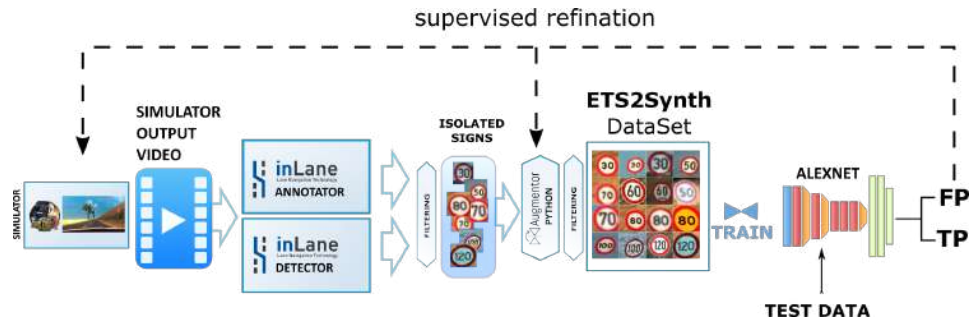

Fig. 2. ETS2Synth dataset generation pipeline. A video sequence is generated from ETS2. Subsequentrly, a labelling stage is performed with an annotator or a detector. Once the dataset is generated, some augmentation processes are applied before a classifier is finally trained. Testing this classifier with real datasets will guide the process, iteratively to incorporate new elements into the scene or new augmentations into the process.

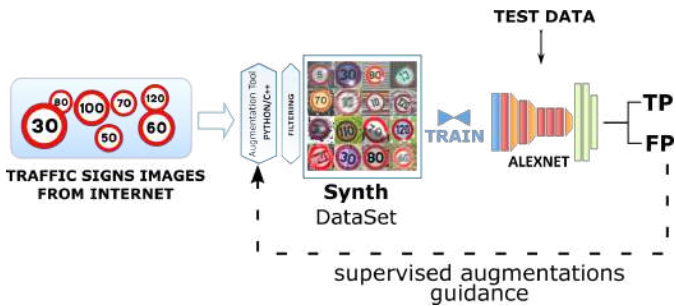

Fig. 3. Synth dataset generation pipeline. Original canonical images are gathered from the internet, an augmentation preprocessing stage is applied to the images, and a classifier is trained. After testing the trained classifier with real datasets, false positives are used to iteratively guide the addition of new augmentations to the generation of the dataset.

not only for ADAS, but also for self-driving vehicles, mobile mapping, and visual road safety inspection. Over the last few decades and before the adoption of DL approaches, several machine learning techniques have been used in TSR [4], [5]. A significant breakthrough was achieved in 2011, when the German traffic sign recognition benchmark (GTSRB) was introduced [6]. Since then, the GTSRB dataset has been adopted as a reference by several authors. In this challenge, traffic sign classification surpassed human performance on GTSRB database using deep learning approaches. In [6], the authors used a committee of neural networks to over-pass human accuracy, reaching $99.15\%$. After the first results of the challenge, several approaches also topped the ranking. Later in 2012, the same team [7] improved recognition performance up to $99.46\%$ by combining various deep neural networks (DNNs) into a multi-column DNN, thus making the system insensitive to variations in contrast and illumination. In [8], twenty independent CNNs with hinge loss function were used to obtain an accuracy of $99.65\%$ on the GTSRB dataset. Until now, the best result on the GTSRB dataset is the one reported in [9], where the authors used a proposal extraction stage guided by an R-CNN-based fully convolutional network and EdgeBox [10] to achieve an accuracy of $99.7\%$. In [11], a new conv-net architecture was introduced to reduce operational cost and maintaining a $99.55\%$ precision. The authors used a lightweight and optimized convnet to detect traffic signs, using LReLU. The model was trained by applying data augmentation processes on the images of the dataset to reduce over-fitting. In [12], the authors probed different network optimization algorithms and combine a model with a spatial transformer network, reaching an accuracy of $99.7\%$ on the GTSRB dataset. In [13], however, the authors focused on mantaining a good accuracy rate of $98.9\%$ while increasing efficiency and reducing the number of parameters and operations when performing inference.

Habibi et al [4], [11] focused on obtaining a lightweight convolutional network to detect and classify traffic signs using the convolutional network layers to simulate a sliding window sampling strategy. This strategy was an attempt to minimize computational cost. Most of these approaches are centered on the stage of recognition and detection. Others, however, focus on the functioning of the overall system. Thus, in [14], the authors proposed a complete system that is able to detect, classify and locate in three dimensions, traffic signs appearing in an image. They combined a two-dimensional single-view traffic signs detection with multi-view scene geometry in an off-line multiple monocular views optimization process. In [15], the authors proposed a complete system to detect, classify, and localize the tridimensional position of the detected signs using RetinaNet for detection and inception for typification, as well as three-dimensional(3D) georeferenced point cloud information obtained by a lidar. Finally, in [16], the authors analyzed the current state of TSR systems and future trends and challenges.

**Synthetic image generation techniques** Training convnets have a drawback of requiring large amounts of training data. This can be solved by investing huge effort, time and money to record, annotate and process video footage from equipped vehicles. Apart from the cost, the recorded media might inadequately represent all the outdoor conditions required to obtain a trained model as generic as possible.

Data augmentation techniques [17], [18] are a popular approach to overcome this drawback as they add variety to the traffic signs data space that models learn. The currently applied techniques are manually designed and they do not always produce the desired results. In [19], the authors state that currently used computational color constancy (a.k.a. white balance) methods do not generate real aspect results; hence, they proposed a novel method to generate more realistic images capable of feeding training processes. Data augmenta-tion techniques have proven highly effective in classification processes [20]. In [21], the authors propose an automatic search for improved data augmentation policies instead of applying randomly different augmentation processes [22]–[24]. Once the augmentation is performed, the output dataset is used to train models to avoid these limitations imposed by the acquisition of real data.

There are also other methods related to the generation of synthetic data, such as [25] where it is proposed to use commercial games output to generate high quality large-scale pixel-accurate ground truth data for training semantic segmentation systems without accessing the game's source code. The work in [26] shows the generation of 3D road scenarios with a previously trained generative model. in [27], a synthetic, generated labeled data was used in the context of crowd counting to propose a fully convolutional network trained with this synthetic data and fine-tuned with real data for crowd counting. Another study analyzed the way of transferring the effects generated by the camera in the acquisition process (chromatic aberration, blur, exposure, noise) to the synthetic data and to propose a novel pipeline [28].

**Training on synthetic data** In [29], the authors analyzed the application of only synthetic data to the training of detection models and concluded that synthetic training data for detection does not produced good results. Further, in the field of detection, authors [30] used a technique called domain randomization to generate synthetic samples to train DNN detectors. This technique entails the randomization of the simulator parameters to generate different scenario conditions. More related to classification field, [31] trained a classifier with synthetically generated images, changing the scope of the image domain to the Russian traffic sign dataset.

In [32] the authors attempted to demonstrate that it is possible to train a neural network using only synthetic data by improving the results obtained with the same network trained with real data. They compared K-NN, LDA, SVM and a CNN with two convolutional and two fully-connected layers. However, they used only two different datasets (GTSRB and STSD) and compared the results of evaluating their testing samples in a classical way. Further, the authors trained classifiers with images obtained from Wikipedia, applying several data augmentation processes to conform the dataset, and using HOG as features to train more traditional classifiers. They concluded that better results are obtained using synthetic data.

In other domains, [33] rendered 1M of synthetic eye images using ray tracing to improve model training, thus incorporating new image information into the model via transfer learning, training with synthetic data first, and then fine tuning with real data.

## III. SYNTHETIC DATASET GENERATION

To minimize the cost of producing and annotating real data, we propose the utilization of synthetic datasets which can be used to train models and in inference time with real data. We explain the proposed method to elaborate a database to train a traffic sign classifier.

Two different sources are used to produce synthetic datasets ( Fig. 2 and 3). On the one hand, a simulator-based approach was used to generate a dataset that we call the ETS2Synth dataset. On the other hand, we used direct augmentation-based generation to create another dataset that we call synth dataset.

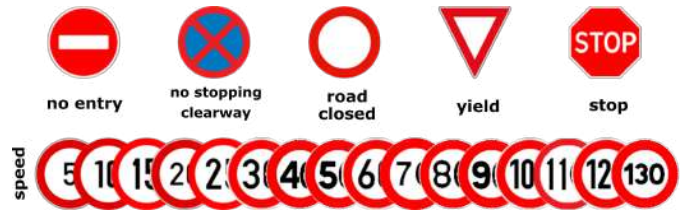

Fig. 4. Selected classes subset (SCS).

In this experiment, a subset of traffic signs classes was selected for the analysis. We call it selected classes subset (SCS)(Fig. 4). This subset was created for efficiency, considering that the classifier behavior will not change when the domain gets larger, as the authors consider that this subset is statistically enough representative.

The SCS is composed of traffic signs with clear visual differences (making it easier to classify) and of another group of speed limitation traffic signs with higher inter-class similarity (making it more difficult to classify).

### A. Simulator-based dataset generation (ETS2Synth Dataset)

As seen in Fig. 2, to obtain the dataset, the process starts generating a video from a simulator sofware engine where the required traffic signs appear, following the pipeline. The acquisition stage was performed by capturing images from a video generated from the renderer view of a simulator. There are some benefits of using simulators instead of simple images:

- It is possible to obtain in a short time, long and dense realistic video sequences of the scenario, thereby allowing the generation of considerably large databases with little effort.
- It is also possible to control the weather, illumination and sensor conditions, thereby producing images that would be otherwise very difficult to obtain.
- In some cases, 3D information is available (e.g., location of the signs) for further processing or automatic labelling (not with our selection).

Nowadays, it is easy to find alternative sources to generate very realistic synthetic data, especially using video games rendering engines. Alternatives in the market are numerous but usually do not offer the possibility to control all the conditions of the road scenarios (weather conditions, sign textures, sign location relative to the camera,etc).

There are several simulators for generating synthetic data, such as EuroTruck Simulator 2 [34], NVIDIA AutoSIM, Prescan [35], Carla [36], AirSim [37], Bus Simulator [38], Grand Theft Auto [39]. In the automotive field, Prescan is one of the most widely used because can define different custom designed scenarios and events, using scripting language. However, to obtain more realistic images we focused on Eurotruck Simulator 2 [34], which is a better rendering engine with numerous scenario-related parameters.

This software also allows the user to create a custom circuit, control lighting, and environmental conditions, and to add traffic signs from different countries. In our case, a simple circuit with various traffic sign was modelled.

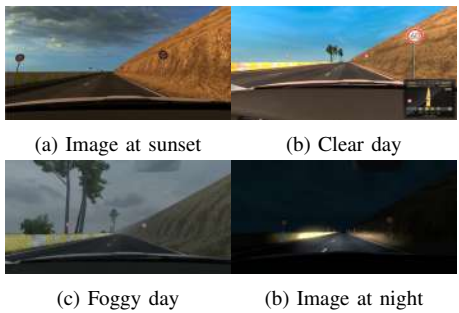

(a) Image at sunset (b) Clear day

(c) Foggy day (b) Image at night

Fig. 5. Images taken from ETS2 simulator custom circuit.

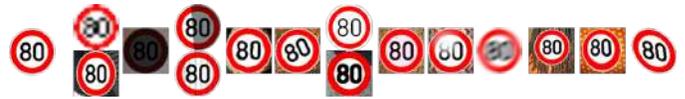

Fig. 6. Examples of transformation processes. From left to right: original, scale, illuminance, shadow, elastic, rotation, morphological operator, noise, specular highlights, blurring, padding, perspective transformation, shearing.

Two hours of footage (sequences of approximately 10 minutes) was generated, including rain (heavy and soft), fog, sun, clouds, and daily and night time conditions (Fig. 5). After the augmentation, the datase contained 17,297 images of speed limitation signs.

Once videos are recorded from the images rendered by the simulator, they are labelled following any of the two proposed ways: automatically (using an existent traffic sign detector) or manually via manual annotation using a labelling tool. For the manual annotation process, we used a video annotation application [40] which has useful tools to speed up the annotation process.

For the automatic alternative, we used an SVM-HOG based detector developed in the context of the H2020 European project INLANE [41]. However, the automatic approach faces a general drawback. Typically, detectors propose not only actual signs but also false positives that the user should filter in a manual validation step. In the case of the INLANE detector, a manual filtering step is added because this detector uses a sliding window sampling strategy and is focused on identifying traffic sign shapes; hence, it generates some false positive samples that must be discarded.

Once the selection is completed, this core dataset is subjected to an augmentation stage. Some of the desired distortions are produced by the simulator. For instance, perspective issues are generated by the simulator as well as climatological conditions or occlusions. Therefore, an intensive augmentation process is unnecessary for simulator generated images. For this task, Augmentor [17], based on python, was used to perform some of the augmentations. It defines a pipeline with the processes to apply and certain probabilities for each process. The output of this augmentation process conforms to ETS2Synth dataset. Finally, a classification model was trained with the output dataset, and tested against a real dataset. For the dataset generation an iteration over the pipeline will be necessary until the performance of the model is adequate. These iterations will guided by the false positive results of the test and will define new traffic signs to incorporate, changes on scenario parameters and new processes over those traffic signs.

### B. Canonical images-based generation (Synth Dataset)

The second approach (Fig.3) entails the application of an intensive augmentation process to a set of canonical images retrieved from the internet [42]–[44]. Compared to the simulator-based approach, there are mainly two advantages of this approach: the time required to generate samples is considerably shorter, and the variance in the data that can be generated via augmentation is higher although not necessarily more realistic.

First, few images per country and type are required to create the canonical set. Then, the augmentation process is an iterative empirical procedure in which several augmentation techniques are applied. After the augmentation processes, the selected model is trained and tested against a real test dataset. This will generate a subset of true positives and another subset of false positives. It is precisely the empirical analysis of this false positives subset that will guide the type of augmentations to be applied in the next iteration, until the desirable results (i.e., a better accuracy) is reached.

### Data augmentation

The process of data augmentation has proven useful in several works [12], [13], [45]. However, it is not straightforward to define a good data augmentation pipeline that generates suitable samples that improve the existing database.

A custom augmentation tool was developed by the authors based on typically applied processes and the addition of new ones (structural shadow and specular highlights addition), in the same pipeline-fashion way as the Augmentor.

Thus, several augmentation processes have been applied to the canonical images: lighting variations, background addition, affine and perspective transformations, blurring, noise addition, histogram normalization, elastic distortion, padding, shadows, specular highlights, and morphological operators. These processes emulate the distortions produced by acquisition devices, and scenario and element conditions, which can distort images in several ways. Some examples of augmentations are illustrated in Fig. 6.

For the ETS2Synth dataset, a light augmentation process was performed to add the variability the simulator could not provide. For the Synth dataset, the augmentation process was considerably deeper. In both cases, a final shallow filtering step on additional generated images was performed to avoid ingesting corrupted or useless samples.

## IV. METHODS

The goal of this study is to determine whether synthetic data can replace real data to reduce the effort and resources dedicated to acquiring of data in deep-learning based solutions.

TABLE I
LIST OF REPRESENTATIVE ONLINE TRAFFIC SIGN DATASETS

| Name | Classes | Country |
|---|---|---|
| Stereopolis (2010) | 10 | FR |
| STSD [46] (2011) | 7 | SW |
| UKOD [47] (2012) | > 100 | UK |
| LISA [48] (2012) | 49 | US |
| GTSRB [49] (2013) | 43 | DE |
| RTSD [50] (2013) | 156 | RU |
| BTS KULD [14] (2014) | > 100 | BE |
| rMASTIF [51] (2015) | 31 | HR |
| TT 100K [52] (2016) | > 45 | CN |
| DITS (2016) | 58 | IT |
| MTSD [53] (2016) | 66 | MY |
| EMTSD (2016) | 66 | MY |
| Mapillary (2018) | 1500 | Global |
| CCTSDB [54](2017) | 48 | CN |
| CURE-TSR [55](2018) | 14 | BE |
| TSRD | 58 | CN |
| VDB | 35 | SP |

ISO 3166-1 alpha-2 codes [56] are used for country codes. Global is used when the origin is from more than one country.

Therefore, a CNN architecture was selected and trained with several real and synthetic datasets and then generated models were compared in a pair-wise mode.

The model selected was AlexNet classification model [24], which won the ImageNet large scale visual recognition challenge in 2012 and has been extensively studied in the literature. We selected a well-known mature model with the objective of avoiding potential flaws in the model's behavior.

When dealing with public datasets, it is necessary to note that they are usually composed of different numbers and types of traffic signs, as seen in Fig. 7. This makes it difficult to compare classifiers that have been trained with different datasets and to draw fair and objective conclusions from the comparison.

In this analysis, we selected six datasets, D = { GTSRB, DITS, rMASTIF, BTSC, TSRD, VDB }, from available public traffic sign datasets (Table I) belonging to different countries.

However, only a subset of the datasets were used (SCS, Fig. 4). The aim of this selection was to run the tests more efficiently based on the assumption that the classes of the SCS were sufficient to extrapolate results to a larger number of classes. Common classes with SCS for each public dataset are described in detail in Fig. 7.

Once the datasets were selected, the next step was the training. In this stage, one model $M_i$ (using AlexNet architecture) for every domain training part $D_i^{TR}$ was generated. However, this training stage could be unnecessary as the model could be previously trained and still be used in comparisons. Several classification experiments were conducted over the signs provided for testing by the selected traffic sign datasets. Thus, a matrix of results was obtained .Each trained model was tested against each dataset (training and testing part). Given that not all the datasets shared the same type of signs, the testing dataset used for each model was formed by the samples belonging to the intersection between the training dataset and the SCS.

From that table ,we extracted the success (true positives, TP)

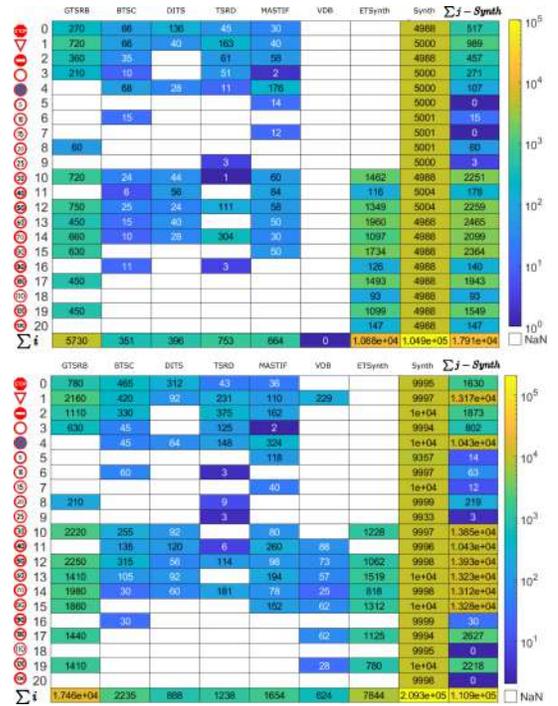

Fig. 7. Number of images per dataset only for classes in the SCS: (top) testing samples, (bottom) training samples.

and failure (false positives, FP) rates for each model against all the considered datasets.

After gathering these results, we produced different statistics to provide additional interpretation to the results: a model vs model correlation matrix ($G$) and the behavior of each model regarding the sign types, i.e., signs vs model matrices ($S$). With these statistics, we could draw objective and fair conclusions on the performance of the classifier under study.

To compare the results between models $M_i$ and $M_j$ and to generate $G$, it is necessary to use a metric. The models were compared in pairs using the precision value obtained. The classes involved in the comparison between two models $M_i$ and $M_j$ are established as $\{C_{i,j}\} = \{C_i\} \cap \{C_j\} \cap \{SCS\}$, where $\{C_i\}$ and $\{C_j\}$ are the set of classes used to train model $M_i$ and $M_j$ respectively.

This metric avoids the influence of uncommon classes between both datasets in the results, which can distort the comparison. Therefore, the models were tested against the same test set $T_{i,j}$ for fairness.

The dataset that will be used to compare the i-th and j-th models, $T_{i,j}$, will be the union of each one of the $T_{i,j}^k$ for each one of the k datasets in the analysis.

$$T_{i,j}^k = \{C_{i,j}\} \cap \{D_k^*\} \quad (1)$$

where $\{D_k^*\}$ is the set of training samples of the dataset $\{D_k^{TR}\}$ or the set of testing samples of the dataset $\{D_k^{TE}\}$, depending on the type of sample set to be analyzed, in testing or training.

$$T_{i,j} = \bigcup_{k=0}^{N} \{T_{i,j}^k\} \quad (2)$$

Here, N is the number of datasets used in the analysis. Once the testing dataset is defined and to calculate each value of the confusion matrix $G_{i,j}$ each model precision, $P_i$ and $P_j$, needs to be calculated. These values will be obtained by testing $T_{i,j}$ against $M_i$ and $M_j$.

$$f : \{T, M\} \to TP, FP$$
$$(TP_p, FP_p) = f(T_{i,j}, M_p) \quad \backslash \quad \forall p \in \{i, j\} \quad (3)$$

where $f$ stands for "inference" and $(TP_p, FP_p)$ are true and false positives of the classifier, respectively; $p \in \{i, j\}$. With these values, precision P is computed, where $P_p = \frac{TP_p}{TP_p + FP_p}$.

In some cases, models are trained with very advantageous datasets which offer few complexities. This favors these models during testing. To avoid this unfairness, both models face the same signs so that the results can be fairly compared. However, computing precisions this way may lead to incorrect interpretations. In this case, some of the dataset cardinalities are much higher than others; hence, joining all the common classes into one large dataset will produce a bias in the results towards those datasets rather than to the smaller ones.

To overcome this problem, we propose a different approach to calculate precision. First, each dataset will be considered separately and will be tested with both models i and j as:

$$\{(TP_p^{i,j,k}, FP_p^{i,j,k}) = f(T_{i,j}^k, M_p)\} \quad (4)$$

where N is the number of datasets, $k \in \{0..N\}$ and $p \in \{i, j\}$, which are the two models under comparison. Therefore, a set of tuples (TP, FP) are computed, one set for each one of the models in the comparison.

$$P_p^{i,j,k} = \frac{TP_p^{i,j,k}}{TP_p^{i,j,k} + FP_p^{i,j,k}} \quad (5)$$

$$P_p = \sqrt[n]{\prod_{k=0}^{n} P_p^{i,j,k}} = \sqrt[n]{\prod_{k=0}^{n} \frac{TP_p^{i,j,k}}{TP_p^{i,j,k} + FP_p^{i,j,k}}} \quad (6)$$

With this approach a set of N precision values ($Pr_i^{i,j,k}$ and $Pr_j^{i,j,k}$) are calculated for $M_i$ and $M_j$. Actual precision for model i and j ($Pr_i$ and $Pr_j$) is obtained as the geometric mean of those N values.

In both cases, the datasets used to compare both models are the same; therefore, they are totally comparable.

Each cell of the model vs model confusion matrix $G_{i,j}$ compares two models which were trained with a different training dataset. The value that appears in the cell is the precision of the model corresponding to that row if that precision value is larger than its complementary one, or the difference (in negative) between the precision values of both models if the value is lower, as seen in Eq. (7).

$$G_{i,j} = \begin{cases} P_i & \text{if } P_i >= P_j \\ P_i - P_j & \text{if } P_i < P_j \end{cases} \quad (7)$$

Other statistics used in this study are illustrated in Figs. 10 and 11. On the one hand, in Fig. 10, $M_i$ is tested against its domain testing dataset $D_i^{TE}$, and on the other hand, in Fig. 11, the testing dataset used is $T_{i,j}$ which is conformed by the union of all the common classes between the SCS and each public testing datasets samples. Fig. 11 relates traffic sign classes with the models in an isolated way. They show the results obtained by each model when tested with each traffic sign class, making it possible to analyze the impact of each traffic sign class in the performance value of the model. Several interesting conclusions which will be discused later, can be retrieved from the second table. Furthermore, these results can guide the process of improving the classifier, pointing out the classes where the classifier could not generalize effectively.

---

**Algorithm 1:** Data Generation algorithm

$T_{i,j}^{k,*}$ =Get From $D_k^*$ all samples which belong to common classes between $D_i^*$, $D_j^*$ and SCS.
$* \in \{TEST, TRAIN\}$;

**GenerateDataTables** $(D^{TR}, D^{TE})$
  **input** : N training and test public datasets.
  **output**: TP and FP rates traffic sign-wise tables for n models trained with each of $D^{TR}$ and tested with $D^{TR}$ and $D^{TE}$
  foreach $M_i \in M$ do
    foreach $D_j \in D$ do
      foreach $* \in \{TR, TE\}$ do
        $DataTable[i][j]^* = Test(M_i, T_{i,j}^{j,*})$;

**ComputePrecision** $(i, j)$
  **input** : first model index, second model index
  **output**: Precision values for $P_i$ and $P_j$
  foreach $D_k \in D$ do
    $TP_i^{i,j,k}, FP_i^{i,j,k}$=Test($M_i$, $T_{i,j}^{k,*}$);
    $TP_j^{i,j,k}, FP_j^{i,j,k}$=Test($M_j$, $T_{i,j}^{k,*}$);
    $\boldsymbol{P_i}^{i,j,k} = TP_i^{i,j,k}/(TP_i^{i,j,k} + FP_i^{i,j,k})$;
    $\boldsymbol{P_j}^{i,j,k} = TP_j^{i,j,k}/(TP_j^{i,j,k} + FP_j^{i,j,k})$;
  $P_i, P_j$=GeometricMean($\boldsymbol{P_i}, \boldsymbol{P_j}$);

**GenerateConfusionMatrix**
  **input** : Data tables
  **output**: Model vs Model matrix (G)
  foreach $M_i \in M$ do
    foreach $M_j \in M$ do
      $P = ComputePrecisions(i, j)$;
      **if** $P_i >= P_j$ **then**
        $G_{i,j} = P_i$;
      **else**
        $G_{i,j} = P_i - P_j$;

---

## V. RESULTS AND DISCUSSION

Several results were obtained from the performed tests: precision tables, model vs model correlation matrix and signs vs model table. Interpreting the data is difficul when heterogeneous datasets are used. First, as seen in Fig. 7, there are several differences between these datasets, not only in the sign types, but also in the number of signs per class, making it more difficult to compare the results. To enable analysts to really ascertain the best model, we propose using heterogeneous testing datasets as we have in this case.

It can also be noticed that several datasets are biased to certain signs; some are more balanced and others insufficient samples. The Synth dataset is the most complete and balanced set of traffic sign class due to its inherent simplicity. The first step in this analysis, was to train AlexNet model with all training datasets. It is important to maintain all the key parameters of the network for all the classifiers to mantain the scenario. Subsequently, each model was tested against all datasets(i.e., both the test and the training samples). The results are shown in [57] for test and training datasets. In these result tables true positives, false positives, and precision are computed for each type of traffic sign in each dataset. This enabled us to subsequently merge the data in the most appropiate way for the analysis. Each cell represents the behavior of the classification process for each traffic sign type in one domain. Marginal values were also computed as they provide good information about the model performance with a particular dataset (columns) or the complexity associated to each sign, i.e., how difficult it is to classify a type of sign (rows).

Some of the values introduce bias in the precision for some of the classes due to the cardinality of signs. Thus, the number of signs of Synth dataset, being so numerous, makes the rest of the results irrelevant, and the average precision of GTSRB turned to 0.583. Therefore, we decided to calculate the performance of the classifier by taking it from the geometric mean of precision percentages. This produced a more balanced result because all the datasets were considered at the same level and the datasets with more samples did not benefit.

Once the table of results was completed for each model, the correlation matrix between the different models was computed to identify the model that best generalizes. Hence, the models were pairwise compared with the same testing set. This testing set was built as explained in the methods section.

## A. Evaluation results between models

We can see from the results (Fig. 8) that the best models are, in order, the ones trained with Synth Dataset, ETS2Synth and GTSRB. This confirms our initial hypothesis that synthetic data can be used to train models that can be applied to real world data. It also confirms that even though the GTSRB model performs better against its own domain (Fig. 10) testing and training datasets than Synth and ETS2Synth datasets, its generalization ability is worse as shown in Fig. 8).

The worst behavior of Synth dataset ocurrs when comparing it to ETS2Synth dataset. Nevertheless, Synth dataset produced better results than ETS2Synth dataset and performed better compared to the other models.

An additional experiment was conducted to incorporate Synth dataset generalization power into a specific domain (Fig. 9). Different combinations were compared with the objective of improving models. This experiment was performed only with GTSRB and Synth datasets. To train new models, the datasets were combined, fine-tuned from already learned models, and trained joining both datasets. These new models were tested against all datasets to obtain cross-domain results that are shown in the figure.

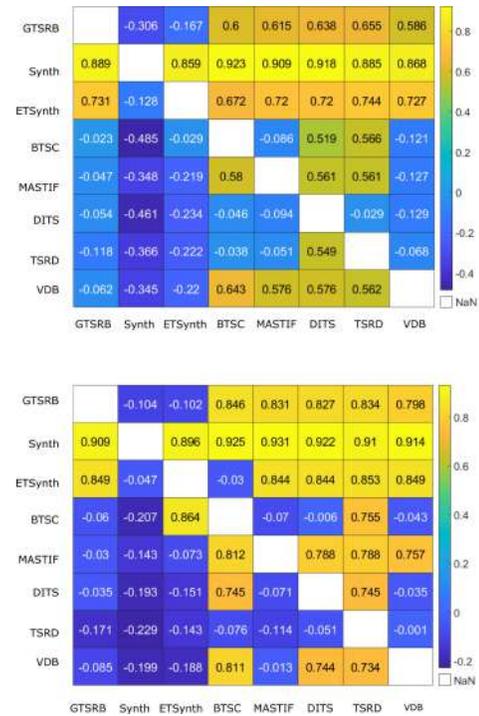

Fig. 8. Comparison between pairs of classifiers using test samples from datasets. Positive means row model performs better than column model.

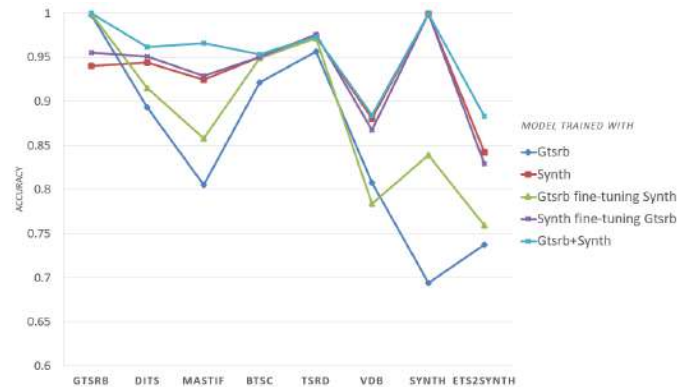

Fig. 9. Comparison between models testing cross-domain.

Finally, to test the behavior of synthetic trained dataset against an extremely challenging real-image dataset, we conducted a final experiment to test CureTSD dataset, which is reputed to be one of the most difficult traffic sign datasets available, with GTSRB-trained, Synth-trained and ETS2Synth-trained classifiers. The results show us that even with very difficult images,synthetic image-based classifiers can compete against real ones with good results. In this case, when comparing GTSRB against Synth dataset, we observe that the results increasingly become very similar both in testing (0.51 for GTSRB , 0.57 for Synth, and 0.76 for ETS2Synth) and training sets ( 0.65 for GTSRB , 0.63 for Synth, and 0.55 for ETS2Synth). It is necessary to consider that Cure-TSD images were acquired in Belgium, which is a very similar domain to the one of GTSRB; this could be beneficial for the GTSRB-trained model, and the Synth-trained model still provides good results. The low values of the performance numbers are noteworthy, and this is due to the presence of low-quality and distorted images in the dataset.

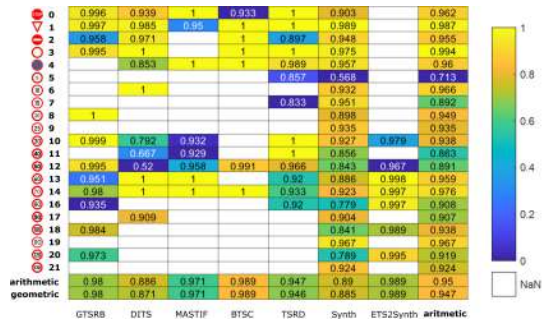

Fig. 10. Results obtained from testing the models against a testing dataset in the same domain. The first column is the model trained with GTSRB dataset training samples against GTSRB dataset testing samples. Heatmap is applied per column to visualize the inter-class variations.

*B. Evaluation results by traffic sign*

Another interesting analysis was performed considering the complexity of classification for each traffic sign type. Not all signs are equally difficult to classify and this depends not only on the type of the sign but also on the training dataset used. This can be seen in Fig 11, where results related to the different signs for each dataset test are presented.

The number of images for each traffic sign inside each dataset must be considered because it can introduce bias in the results. It can be seen in 10 that Synth dataset does not provide the best accuracy when tested against its own dataset. GTSRB, for instance, produces 0.98 of accuracy attending to the geometric mean while the Synth-trained model produces a 0.88. Thus, although the result against the domain testing dataset is good, it does not mean the model's generalization power will be better than another with worse results.

It is also remarkable that there is not a single most difficult class for all classifiers; rather, each model has its own difficult classes. For example, GTSRB presents difficulties when classifying 100 speed limit signs, whereas and Synth encounters more problematic 5 speed limit signs or no entry signs. In the synthetic dataset generation stage, these classes can cause modifications in the applied processes that will improve the dataset used to train classification model.

## VI. CONCLUSIONS

Typically, a classifier is evaluated against the corresponding test samples provided by the dataset. The results may be biased because trained models learn the peculiarities of the domain, such as perspective transformations or image distortions produced by the type of cameras used. Consequentrly, they usually perform very well against their own test dataset.

In this paper, we discussed how a synthetic dataset performs almost as well as a model trained with real data if suitable preprocesses are applied to the synthetic images to prevent a bias against the peculiarities of real dataset.

Both hypotheses were validated in our analysis. On the one hand, both Synth and ETS2Synth trained models performed better (reaching precision values approximately 90%) in generalization models trained using real datasets, as seen in Fig. 8.

Therefore, if the target domain is general, the best option is to create a synthetic dataset trained model, as it has the best behavior compared to the other models( Fig. 8). However, if the target domain is very focused, a good alternative could be to use a model trained with images of that domain, while a better alternative ( Fig. 9) is to mix both real and synthetic datasets to train a model. This is the best choice to keep a good performance against unseen domains.

On the other hand, the training process can be guided considering the signs vs models analysis. We used this approach and the results improved iteratively. The Synth and ETS2Synth dataset based models have the best behavior in generalizing. Here, we can observed the bias produced by learning with only one dataset as the synthetic-based model produced worse results than GTSRB or DITS models in its own test (Fig. 10). This is because the variability in the synthetic dataset is very large; thus, training and testing sets differ significantly, compared to the situation of a real dataset. The way to proceed is to continue detecting the most difficult signs for Synth or ETS2Synth model and generating new samples by empirically choosing the augmentation process to apply to canonical samples. To iterate, training starts from the last learnt model, as a transfer learning process between iterations of the model, or by freezing the training steps for some neural network paths that are already good enough to force the model to improve the classification of other signs.

Synthetic-based models can achieve good precision results when applied to real data. Furthermore, the process to obtain this data is, significantly less expensive than obtaining real images. In addition, the specificities or peculiarities of the real data typically decrease the generalization capabilities of the models.

It is also interesting to note the possibility of applying this kind of analysis to the field of detection, generating images using samples and background images, and merging them in a random way using all the augmentation utilized in this paper.

In addition, to validate the conclusions reached, we extended the analysis by testing the models against the training samples of the datasets, which are significantly larger than the testing sets. These results can be consulted at the referenced website.

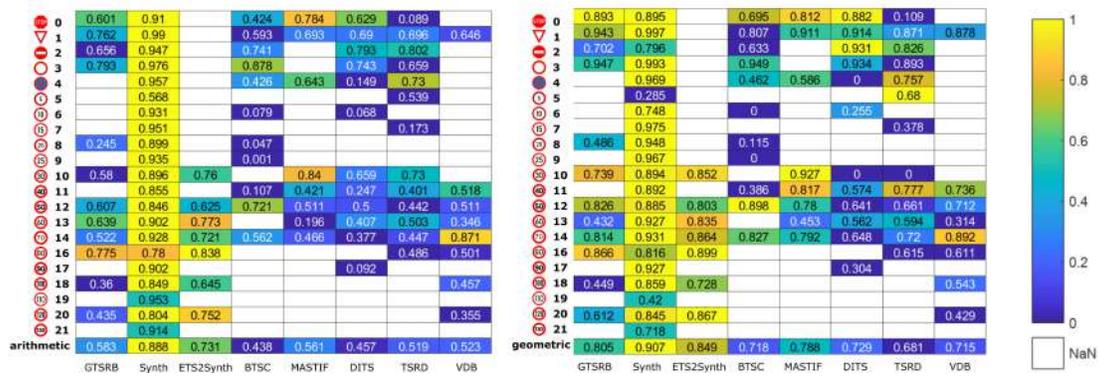

Fig. 11. Results obtained from testing models against common testing dataset from the intersection of common classes between all datasets and the subset of signs of interest. Heatmap is applied per rows. The left part was calculated using aritmetic mean and right part was calculated using geometric mean.

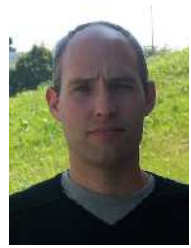

**Andoni Cortés** received M.S. degree in Computer Science in 2001 from the University of the Basque Country (FISS), Spain. He is currently (since 2007) working as a researcher at the the Intelligent Transport Systems and Engineering Area of Vicomtech. He has been involved, as technical and scientific coordinator, in local projects mainly focused on applying computer vision technologies to innovative industry and transportation solutions. He has also participated as a researcher in some European cutting-edge projects (e.g., InLane). His research interests are in the field of computer vision, mainly in image processing algorithms and machine learning techniques.

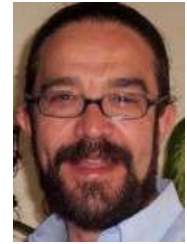

**Pr. Clemente Rodríguez** received a Bachelor's degree in Computer Science from the Autonomous University of Barcelona, Spain in 1981 and a degree in Computer Science in 1986 from the Polytechnic University of Catalonia, Spain. He is a Professor at the Computer Architecture and Technology Department of the Basque Country University where he has been working since 1988. Since 1990, he has lead a group in parallel and computer architectures which has been participating in several European and local research projects. Since 2006 he has been a member of the Computer Architecture Group of the University of Zaragoza. His interests include timing analysis and memory hierarchy.

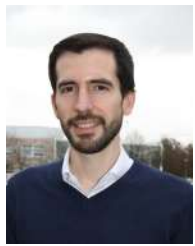

**Dr. Gorka Vélez** received an M.Sc. degree in Electronic Engineering from the University of Mondragon (Spain) in 2007, and a Ph.D. from the University of Navarra (Spain) in 2012. He currently works as a Senior Researcher at the Intelligent Transportation Systems (ITS) and Engineering Department of Vicomtech (Spain). His research is focused on applying machine learning technologies on the ITS and industrial sectors. He was the technical coordinator of the H2020 project INLANE, and he is currently involved in TransSec and 5G-MOBIX.

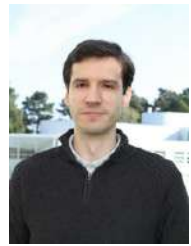

**Mr. Javier Barandiarán** received an M.S. degree in Computer Science from the University of the Basque Country, Spain, in 2004. He is currently pursuing a Ph.D. He has been a senior researcher at Intelligent Transport Systems and Engineering Area of Vicomtech since 2007. His research interests are in the field of computer vision, augmented reality, tracking and 3D reconstruction.

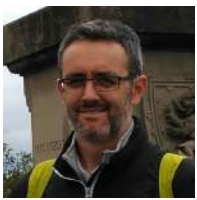

**Dr. Marcos Nieto** received M.S. and Ph.D. degree in Electrical Engineering from ETSIT of the Universidad Politécnica de Madrid (UPM), Spain, in 2005 and 2010, respectively. From 2005 to 2010 he worked as Researcher within the Image Processing Group at the UPM. Since 2010 he has worked as a Researcher on computer vision and machine learning techniques of the Intelligent Transport Systems and Engineering Area of Vicomtech. He has been the technical and scientific coordinator of FP7 and H2020 projects and has experience in transferring technology to industry to support of innovation. He is the author of more than 60 peer reviewed international publications in relevant conferences (+40) and journals (+15). He is also an active reviewer of prestigious journals for IEEE, Elsevier and Springer.